\documentclass[10pt,twocolumn,letterpaper]{article}

\usepackage{cvpr}              % To produce the CAMERA-READY version
\usepackage{graphicx}
\usepackage{amsmath}
\usepackage{amssymb}
\usepackage{booktabs}
\usepackage{multirow}
\usepackage[dvipsnames]{xcolor}
\usepackage{enumitem}
\usepackage{microtype}
\usepackage{nicematrix}

\usepackage[pagebackref,breaklinks,colorlinks,linkcolor=blue,citecolor=blue]{hyperref}

\newcommand{\Ourstight}{NeurOCS}
\newcommand{\Ours}{NeurOCS }

\newenvironment{packed_lefty_item}{
\begin{itemize}[leftmargin=*]
\vspace{-6pt}
  \setlength{\itemsep}{0pt}
  \setlength{\parskip}{0pt}
  \setlength{\parsep}{0pt}
  \setlength{\topsep}{-10pt}
  \setlength{\partopsep}{0pt}
}{\end{itemize}\vspace{-6pt}}

\usepackage[capitalize]{cleveref}
\crefname{section}{Sec.}{Secs.}
\Crefname{section}{Section}{Sections}
\Crefname{table}{Table}{Tables}
\crefname{table}{Tab.}{Tabs.}

\def\cvprPaperID{2104} % *** Enter the CVPR Paper ID here
\def\confName{CVPR}
\def\confYear{2023}

\begin{document}

%\title{Exploiting a Good Shape Supervision for Object Localization}
%\title{\Ours: Learning from Neural NOCS Supervision for \\ Monocular 3D Object Localization} 
\title{NeurOCS: Neural NOCS Supervision for Monocular 3D Object Localization}
% visual scope?
% CADs does not matter: Learning from Neural NOCs Supervision for 3D Object Localization

\author{Zhixiang Min$^1$ \quad Bingbing Zhuang$^2$ \quad Samuel Schulter$^2$ \quad Buyu Liu$^2$ \\
Enrique Dunn$^1$ \quad Manmohan Chandraker$^2$  \vspace{+0.3em} \\
$^1$Stevens Institute of Technology~~~$^2$NEC Laboratories America\vspace{-0em} \\
}

\maketitle

%%%%%%%%% ABSTRACT
\begin{abstract}
Monocular 3D object localization in driving scenes is a crucial task, but challenging due to its ill-posed nature. Estimating 3D coordinates for each pixel on the object surface holds great potential as it provides dense 2D-3D geometric constraints for the underlying PnP problem. However, high-quality ground truth supervision is not available in driving scenes due to sparsity and various artifacts of Lidar data, as well as the practical infeasibility of collecting per-instance CAD models. In this work, we present \Ourstight, a framework that uses instance masks and 3D boxes as input to learn 3D object shapes by means of differentiable rendering, which further serves as supervision for learning dense object coordinates. 
Our approach rests on insights in learning a category-level shape prior directly from real driving scenes, while properly handling single-view ambiguities. Furthermore, we study and make critical design choices to learn object coordinates more effectively from an object-centric view. Altogether, our framework leads to new state-of-the-art in monocular 3D localization that ranks 1st on the KITTI-Object~\cite{geiger2012we} benchmark among published monocular methods.
\end{abstract}

\section{Introduction}
\label{sec:intro}

\begin{figure}[t!]
    \centering
    \includegraphics[width=1\linewidth]{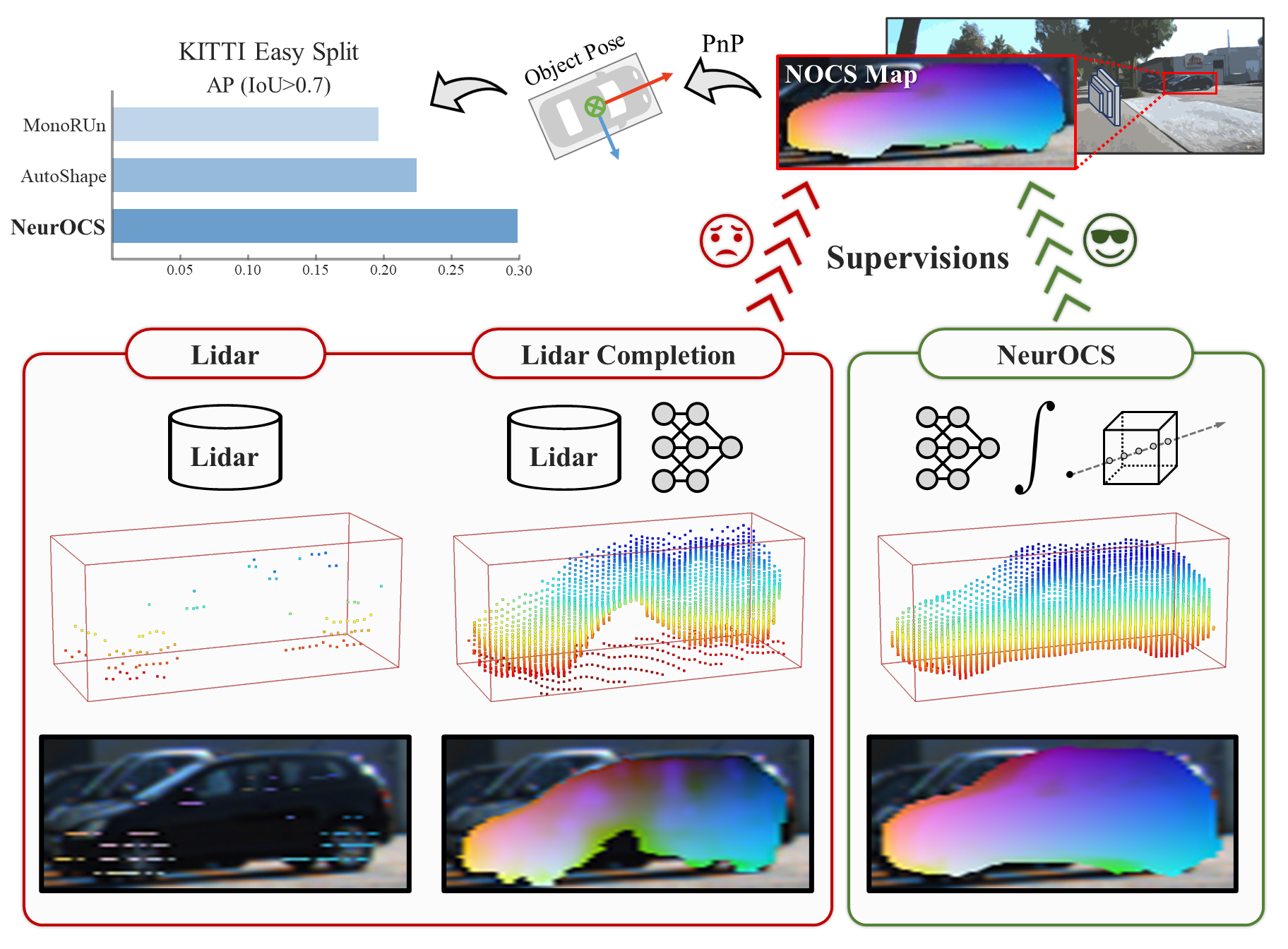}
    \vspace{-1.8em}
    \caption{\textbf{\Ourstight} learns category-level shape model in real driving scenes to provide dense and clean NOCS supervision through differentiable rendering,
    leading to new state-of-the-art 3D object localization performance.
    }
    \label{fig:teaser}
    \vspace{-0.5em}
\end{figure}

Localization of surrounding vehicles in 3D space is an important problem in autonomous driving. While Lidar \cite{wu2022sparse,Shi_2019_CVPR,lang2019pointpillars} and stereo \cite{wu2022sparse,Shi_2019_CVPR,lang2019pointpillars} methods have achieved strong performances, monocular 3D localization remains a challenge despite recent progress in the field \cite{ma2021delving,brazil2020kinematic,li2022densely}.

Monocular 3D object localization can be viewed as a form of 3D reconstruction, with the goal to estimate the 3D extent of an object from single images. While single-view 3D reconstruction is challenging due to its ill-posed nature, learned priors combined with differentiable rendering \cite{tewari2022advances} have recently emerged as a powerful technique, which has the potential to improve 3D localization as well. Indeed, researchers~\cite{Kundu_2018_CVPR,beker2020monocular,zakharov2020autolabeling,zakharov2021single} have applied it as a means for object pose optimization in 3D localization. 
%However, their accuracy are currently far from leading methods in standard benchmark~\cite{geiger2012we}, largely 
However, difficulties remain due to pose optimization being ambiguous under challenging photometric conditions (such as textureless vehicle surfaces) and geometric occlusions ubiquitous in driving scenes. The question of how differentiable rendering may be best explored in 3D localization remains under-studied. Our work proposes a framework to unleash its potential -- instead of using differentiable rendering for pose optimization, we use it with annotated ground truth pose to provide high-quality supervision for image-based shape learning, leading to a new state-of-the-art in 3D localization performance. 

Our framework relies on the machinery of Perspective-n-Point (PnP)~\cite{lepetit2009epnp} pose estimation, which uses 2D-3D constraints to explicitly leverage geometric principles that lend itself well to generalization. In particular, learning 3D object coordinates for every visible pixel on the object surface, known as normalized object coordinate space (NOCS)~\cite{wang2019normalized}, provides a dense set of constraints.
However, despite NOCS-based pose estimation dominating indoor benchmarks~\cite{hodan2020bop}, their use in real driving scenes has been limited primarily due to lack of supervision -- it is nontrivial to obtain accurate per-instance reconstructions or CAD models in road scenes. Lidar or its dense completion~\cite{hu2021penet} are natural alternatives as pseudo ground truth~\cite{chen2021monorun}, but they are increasingly sparse or noisy on distant objects with reflective surfaces. Synthetic data is a potential source of supervision \cite{zakharov2020autolabeling,zakharov2021single}, but is inherently restricted by domain gap to real scenes.

In this work, we propose \textit{\Ours} that leverages neural rendering to obtain effective NOCS supervision. Firstly, we propose to learn category-level shape reconstruction from real driving scenes with object masks and 3D object boxes using Neural Radiance Field (NeRF) \cite{mildenhall2021nerf}. The shape reconstruction is then rendered into NOCS maps that serve as the pseudo ground-truth for a network dedicated to regress NOCS from images. Specifically, \Ours learns category-level shape representation as a latent grid~\cite{tiny_cuda_nn,mueller2022instant} with low-rank structure, consisting of a canonical latent grid plus several deformation bases to account for instance variations. With single-view ambiguities handled by a KL regularization~\cite{kingma2013auto} and dense shape prior, we show that the NOCS supervision so obtained yields strong 3D localization performance, even when the shape model is trained without using Lidar data or any CAD models. Our NOCS supervision is illustrated in \cref{fig:teaser} in comparison with Lidar and its dense completion. We also note that NeRF rendering is only required during training, without adding computational overhead to inference. We show \Ours is complementary to direct 3D box regression~\cite{ma2021delving,peng2022did}, and their fusion further boosts the performance.

Further, we study crucial design choices in image-conditioned NOCS regression. For example, as opposed to learning NOCS in a scene-centric manner with the full image as network input, we learn in an object-centric view by cropping objects without scene context, which is demonstrated to especially benefit the localization of distant or occluded objects. Our extensive experiments study key choices that enable \Ours to achieve top-ranked accuracy for monocular 3D localization on the KITTI-Object benchmark \cite{geiger2012we}.

\noindent In summary, our contributions include:
\begin{packed_lefty_item}
\item  We propose a framework to obtain neural NOCS supervision through differentiable rendering of the category-level shape representation learned in real driving scenes.

\item We drive the learning with deformable shape representation as latent grids with careful regularizations, as well as 
 effective NOCS learning from an object-centric view.  
\item Our insights translate to state-of-the-art performance which achieves a top rank in KITTI benchmark~\cite{geiger2012we}.
\end{packed_lefty_item}

\begin{figure*}[!t]
    \centering
    \vspace{-0.5em}
    \includegraphics[width=0.95\linewidth]{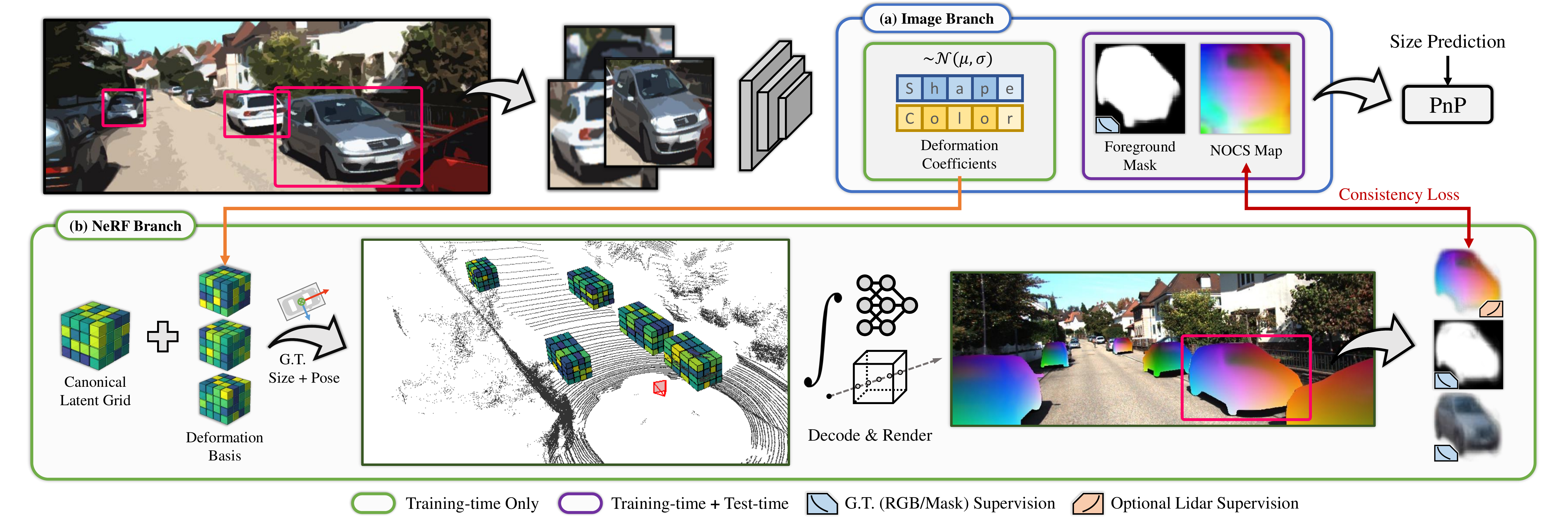}
    \vspace{-0.5em}
    \caption{\textbf{Overview of \Ourstight.} 
    Given each detected object, our network predicts the object mask, NOCS map, and its deformation coefficients associated with a categorical NeRF model. The jointly trained NeRF renders the NOCS supervision for the network prediction branch. During inference, only the predicted object mask and NOCS map are consumed by PnP for localization. 
    }
    \label{fig:workflow}
    \vspace{-0.5em}
\end{figure*}

\section{Related Work}
\label{sec:relatedwork}
\vspace{0.1cm}
\noindent \textbf{Direct regression methods.} These methods directly regress 3D bounding boxes parameters~\cite{simonelli2019disentangling,brazil2019m3d,chen2016monocular,mousavian20173d,zhang2022dimension,ye2022consistency}.
In light of depth being the most critical factor~\cite{ma2021delving,jing2022depth}, many works develop  
\textit{depth}-guided localization, including depth-aware networks~\cite{brazil2019m3d,Ding_2020_CVPR_Workshops,kumar2022deviant,huang2022monodtr}, depth-conditioned message propagation~\cite{wang2021depth}, depth-equivariant network~\cite{kumar2022deviant}, depth-guided feature projection~\cite{reading2021categorical}, depth-based feature enhancement~\cite{bao2019monofenet}, depth pretraining for knowledge distillation~\cite{park2021pseudo}, depth from motion~\cite{wang2022monocular}, and object depth decomposition~\cite{park2021pseudo} for affine data augmentation.
In addition, \cite{qin2022monoground,liu2021ground,chen2016monocular,gu2022homography,zhou2021monocular,mogde2022,min2020voldor,min2021voldor+} utilize the ground plane as depth prior for localization.   Researchers~\cite{wang2019pseudo,you2019pseudo,qian2020end,ma2020rethinking,simonelli2021we,wang2021progressive} also convert depth maps into the pseudo-Lidar representation
to directly apply advanced Lidar-based methods. \cite{hong2022cross,yang2022mix,peng2022lidar} utilize pseudo labels or teacher-student training to learn from unlabeled data.

%\cite{chen2022pseudo}
%\cite{lian2022exploring}
%\cite{zou2021devil}

\vspace{0.1cm}
\noindent \textbf{Geometric reasoning methods.} These methods~\cite{mousavian20173d,Ku_2019_CVPR} solve object poses with 2D-3D perspective constraints, such as box edge correspondences~\cite{mousavian20173d}, object heights~\cite{shi2021geometry,lu2021geometry,kumar2022deviant},
sparse keypoints~\cite{li2020rtm3d,chabot2017deep,liu2021autoshape}, or edges~\cite{li2022densely} on the object surface. EPro-PnP~\cite{chen2022epro} learns localization-friendly correspondences without explicit shape constraints, but it has limited performance in KITTI that has relatively small amount of training data. Notably, a line of methods learn per-pixel object coordinates known as NOCS~\cite{wang2019normalized} to establish dense constraints. In view of the noisy nature of Lidar, MonoRUn~\cite{chen2021monorun} proposes a self-supervised way to learn NOCS by minimizing reprojection error, which nonetheless does not provide strong shape constraints and still requires Lidar to achieve good accuracy.
Some methods~\cite{zakharov2020autolabeling,zakharov2021single} resort to synthetic data for NOCS supervision, which suffers from domain gap.  In this work, we show that NOCS can be learned effectively from high-quality NeRF-rendered supervision. This also draws connection to  AutoShape~\cite{liu2021autoshape} as an auto-labeling pipeline that uses CAD models to generate pseudo ground truth for sparse keypoints. Our method distinguishes itself by directly learning dense NOCS from real driving scenes and working well even without using CAD models and Lidar supervision. Also worth noting is MonoJSG~\cite{lian2022monojsg} that leverages learned NOCS to perform cost volume based object depth refinement, further validating the merits of NOCS.

%\cite{wang2022monocular}

\vspace{0.1cm}
\noindent \textbf{Hybrid methods.} Regression-based and geometry-based methods are not mutually exclusive but rather complementary. \cite{zhang2021objects,lu2021geometry,kumar2022deviant} combine depth from regression and depth from height for improved accuracy. MonoDDE~\cite{li2022diversity} further enriches the set of depth cues and fuses them in an end-to-end trainable framework. While focusing on NOCS-based geometric reasoning, our method when fused with direct depth regression results in improvements as well  (\cref{sec:kittibenchmak}), demonstrating their desirable complementary nature. 

\vspace{0.1cm}
\noindent \textbf{Differentiable rendering.} 
A number of works~\cite{Kundu_2018_CVPR,beker2020monocular,zakharov2020autolabeling,zakharov2021single, Min_2022_CVPR} have applied differentiable rendering to the 3D localization problem. They typically use it as a means to optimize object pose by minimizing photometric or feature-metric error through a render-and-compare manner. However, pose optimization may be highly ambiguous under the challenging conditions (\cref{sec:intro}) in real driving scenes, especially with a single view. Hence these methods do not currently dominate the standard 3D localization benchmark~\cite{geiger2012we}. In this work, we use differentiable rendering as well but with annotated object poses, and obtain NOCS supervision that leads to state-of-the-art performance in 3D localization.
We also note that the recent success of NeRF~\cite{mildenhall2021nerf} continues to promote differentiable rendering research~\cite{muller2022autorf,Kundu_2022_CVPR,ost2021neural} in driving scenes. However, they assume known 3D object boxes as input, with a focus solely on rendering.

\section{Method}
\subsection{Overview}
 
\paragraph{Problem Formulation.} 
We aim to localize each object as an enclosing 3D box parameterized by its 3D dimension $\mathbf{s}\!=\![l,h,w]$ and the 4-DoF pose parameters 
%$[\theta,x,y,z]$, 
including the yaw-angle $\theta$ and object center $\mathbf{t}\!=\![x,y,z]$. 
Our work studies the NOCS-based approach and solves the pose as a PnP problem. 
Specifically, we denote the position of each 3D point on the object surface as $\mathbf{x}_i \!=\! [x_i^{[x]}, x_i^{[y]}, x_i^{[z]}]^T$ represented in its object coordinate system.  $\mathbf{x}_i$ is further decoupled as the scale-invariant coordinate $\mathbf{o}_i$ in normalized object coordinate space (NOCS) times the object size $\mathbf{s}_i$, i.e. $\mathbf{x}_i \!=\! \mathbf{o}_i \odot \mathbf{s}_i$, for effective learning.  
With network-predicted $\mathbf{o}_i$ and $\mathbf{s}_i$ on a pixel with normalized camera coordinate $\mathbf{p}_i \!=\! [u_i, v_i, 1]^T$,
we solve the pose by minimizing the reprojection error as
\begin{equation} \label{eq:pnp_loss}
    \underset{\mathbf{R}_{\theta}, \mathbf{t}}{argmin} \sum_i \rho \Big( w_i  \big(
    \frac{\mathbf{R}_{\theta} \mathbf{x}_i + \mathbf{t}}{[\mathbf{R}_{\theta} \mathbf{x}_i + \mathbf{t}]_z} - \mathbf{p}_i \big) \Big),
\end{equation}
where $[\,\cdot\,]_z$ denotes the z-axis component, $\mathbf{R}_{\theta}$ is the rotation matrix 
form of the yaw $\theta$,
$w_i$ is the confidence weight for each prediction and $\rho$ denotes a robust M-estimator, where we use Huber loss throughout this work.

\vspace{0.1cm}
\noindent \textbf{Framework Overview.}  
\cref{fig:workflow} shows an overview of \Ourstight's key components. \Ours predicts NOCS as input to PnP estimation of object pose, with a NeRF used to render pseudo-ground truth NOCS for supervision.
Specifically, we first use a separately trained base 3D detector~\cite{ma2021delving,peng2022did} to obtain 2D detections with their 3D size predictions.
We then crop each detected object and use a ResNet50~\cite{he2016deep} followed by a few regression heads to predict its object mask, NOCS map, as well as two sets of coefficients representing the shape and color of the object instance.
The coefficients account for instance variations and are used for deforming a NeRF-based shape model, represented by latent grids as detailed in \cref{sec:nerf} and regularized with deformation and shape priors discussed in \cref{sec:regularization}. 
We jointly train the NeRF and the prediction network -- the NeRF is trained with the ground-truth \textit{3D boxes} (i.e. pose and size), \textit{object mask}, \textit{RGB color}, and optionally \textit{Lidar} data; and the prediction network is trained with the ground-truth \textit{object mask} and the NOCS map rendered from NeRF.
During inference, NeRF is switched off and only the predicted NOCS map and object mask are consumed by PnP for pose estimation (\cref{sec:localization}), where we also predict a confidence score. The PnP solution may be fused with the complementary direct depth prediction additionally regressed by the base 3D object detector. 
Note that our prediction network is decoupled from the base 3D detector, for reasons detailed in \cref{sec:keydesignchoice}.  

\subsection{Categorical Shape Model} \label{sec:nerf}

\noindent\textbf{Shape Representation.} Inspired by InstantNeRF~\cite{mueller2022instant}, we employ a 3D latent grid $\phi$ as our shape representation owing to its efficiency in training and inference.
In particular, we use a multi-resolution dense grid as in \cite{tiny_cuda_nn} to model shape in a unit cube.
For an input NOCS point $\mathbf{o}_i$, we  query the grid by trilinear interpolation and stack the multi-resolution outputs to return a $D$ dimensional feature vector, which can be efficiently decoded by a small MLP network to output density and RGB color.  Our shape representation does not model the view-dependent effects as in vanilla NeRF.

\vspace{0.15cm}
\noindent\textbf{Deformation.} To model the categorical shape variations, we use a set of learnable 3D latent grids to compose a low-rank deformable shape representation. We define a canonical latent grid $\phi_\mu$ and a set of deformation grid basis $\{\phi_i \;|\; i=1 \cdots B\}$ where $B$ is the number of bases. 
For each object, given a coefficient $\mathbf{z} \!\in\! \mathbb{R}^B$ predicted by network, we construct an instance-specific latent grid $\Phi$ by deforming $\phi_{\mu}$ with the linearly combined bases:
\begin{equation} \label{eq:phi_obj}
    \Phi = \phi_\mu + \frac{\sum_{i=1}^B z_i\phi_i}{B}.
\end{equation}
We maintain separate latent grids and coefficients for shape and color. 
By sharing the small number ($<\!<$ grid dimension) of bases, the latent grid is in a low-rank space that forces the deformation to explore categorical commonalities in shape.

\vspace{0.15cm}
\noindent\textbf{Object Volume Rendering.} %This grid is compatible with volume rendering. 
For each object, we use the ground-truth 3D box (i.e. size and pose) to transform the viewing ray into its normalized object coordinate, denoted as
$\mathbf{r}^{[\gamma]} \!=\!  (\mathbf{q}~\!+\!~\gamma \mathbf{d}) \circ \frac{1}{\mathbf{s}}$, where $\mathbf{q}$, $\mathbf{d}$ and $\gamma$ respectively denote the camera center, viewing direction and distance along the ray.
The color $\mathbf{c}(\mathbf{r})$ can be rendered following~\cite{mildenhall2021nerf} as 
\begin{equation}
%\resizebox{0.9\columnwidth}{!}{$
\begin{gathered}
    \mathbf{c}(\mathbf{r}) = \int_{\gamma_{n}}^{\gamma_{f}} \alpha(\mathbf{r}^{[\gamma]}) \, \Phi^{[\sigma]}\big(\mathbf{r}^{[\gamma]}\big) \, \Phi^{[\mathbf{c}]}\big(\mathbf{r}^{[\gamma]}\big) \, d\gamma,
    \\
    and \quad \alpha(\mathbf{r}^{[\gamma]}) = exp\Big(\!- \!\! \int_{\gamma_n}^\gamma \Phi^{[\sigma]}\big(\mathbf{r}^{[\zeta]}\big) d\zeta \Big),
\end{gathered}
%$}
\end{equation}
\noindent where $\Phi^{[\mathbf{c}]}(.)\!\in\! {\mathbb{R}^3}$ and $\Phi^{[\sigma]}(.)\!\in\! {\mathbb{R}^1}$ denote decoding the latent grid $\Phi$ at a given query point into color and density, respectively.
The near and far distance $\gamma_n$ and $\gamma_f$ are given by the ray intersection with the 3D box; therein points are sampled as illustrated in \cref{fig:workflow}.
$\alpha(\mathbf{r}^{[\gamma]})$ is the accumulated transmittance along the ray. The occupancy map indicating object mask is rendered by \begin{equation} \label{eq:render_occ}
    \mathbf{m}(\mathbf{r}) = {\int_{\gamma_{n}}^{\gamma_{f}} \alpha(\mathbf{r}^{[\gamma]}) \, \Phi^{[\sigma]}\big(\mathbf{r}^{[\gamma]}\big) \, d\gamma}.
\end{equation}
%With a slight modification, 
We render the NOCS map by integrating the NOCS as
\begin{equation} \label{eq:render_noc}
    \mathbf{o}(\mathbf{r}) = \frac{\int_{\gamma_{n}}^{\gamma_{f}} \alpha(\mathbf{r}^{[\gamma]}) \, \Phi^{[\sigma]}\big(\mathbf{r}^{[\gamma]}\big) \, \mathbf{r}^{[\gamma]} \, d\gamma} 
    {\mathbf{m}(\mathbf{r})}.
\end{equation}

\vspace{0.15cm}
\noindent\textbf{Shape Losses.} Our shape model is trained by several $L_2$ losses including occupancy loss, color (RGB) loss and additional NOCS supervision from lidar and its  completion~\cite{hu2021penet},
\begin{equation} \label{eq:nerf_losses}
    \mathcal{L}_{shape} = \mathcal{L}_{occ} + \mathcal{L}_{rgb} + \underbrace{\big( \mathcal{L}_{lidar} + \mathcal{L}_{licomp} \big)}_{optional}.
\end{equation}
The occupancy loss is supervised by the ground-truth mask containing 3 categories including foreground, background and unknown (usually due to occlusion), similarly to \cite{muller2022autorf}. We enforce the occupancy to be 1 at foreground, 0 at background and skip the unknowns (see supplementary for examples).
$\mathcal{L}_{rgb}$, $\mathcal{L}_{lidar}$, and  $\mathcal{L}_{licomp}$ are only applied at foreground regions.
For $\mathcal{L}_{lidar}$ and $\mathcal{L}_{licomp}$, we convert point clouds inside the 3D box into NOCS and supervise the corresponding pixels.
While being an indirect shape regularization,
we found the photometric constraints from $\mathcal{L}_{rgb}$ benefit the performance.
For training the latent bases, we only select high-quality examples that are at least $k$ pixels in height ($k = 40$) and have no occlusions. Otherwise, we freeze the latent bases and only optimize their coefficient predictions.

\if false
\begin{figure}[!t]
    \centering
    \vspace{-0.5em}
    \includegraphics[width=\columnwidth]{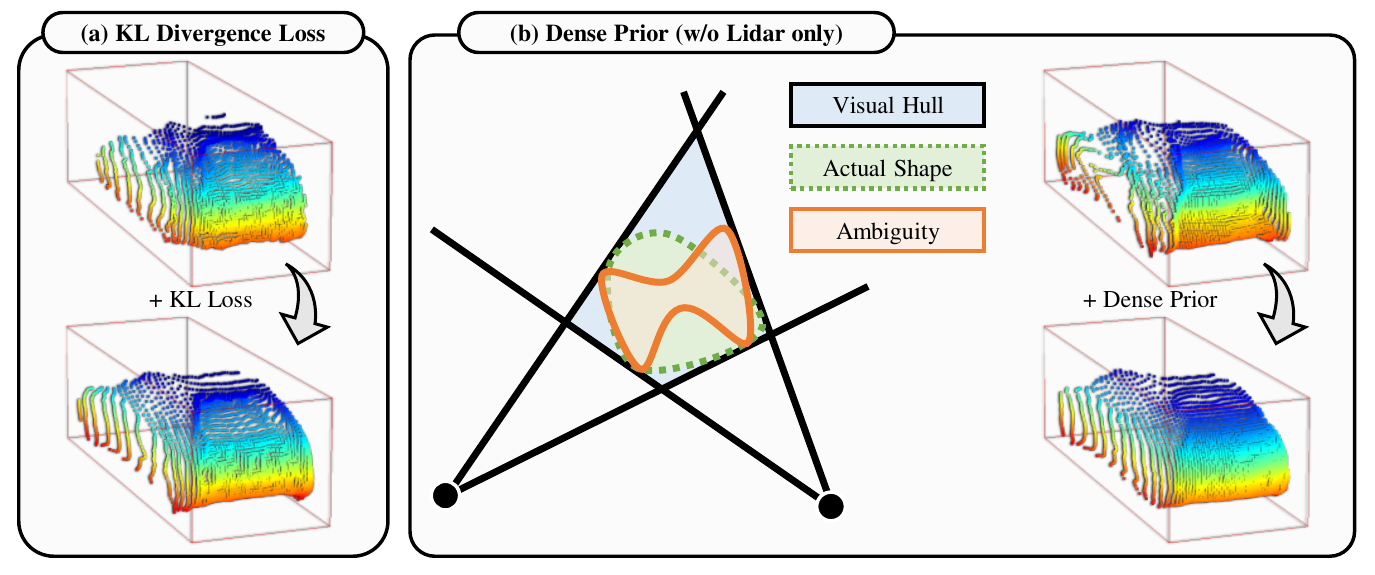}
    \caption{\textbf{Shape Regularizations.} (a) An actual example showing the KL loss yields cleaner shape. 
    (b) 
     The visual hull ambiguity addressed by the dense prior,  shown on an actual example.
    }
    \label{fig:visual_hull_ambiguity}
    \vspace{-1.0em}
\end{figure}
\fi

\subsection{Categorical Shape Regularization}
\label{sec:regularization}
\noindent\textbf{Deformation Regularization.} 
Despite the shared low-rank deformation bases across objects,
their deformations may still not be well-regularized under an ill-posed single-view reconstruction.
Hence, we predict $\mathbf{z}$ as a learnable Gaussian distribution
$q(\mathbf{z} \,|\, \mathbf{I})$ and
add the KL divergence loss \cite{kingma2013auto} as a regularization 
%the coefficient distribution  $q(\mathbf{z} \,|\, \mathbf{I})$ 
minimizing  information gain 
as in VAEs~\cite{kingma2013auto}, 
\begin{equation}
    \mathcal{L}_{kl} = KL \big( q(\mathbf{z} \,|\, \mathbf{I}) \;||\; p(\mathbf{z}) \big),
\end{equation}
\noindent where $p(\mathbf{z})\!\sim\! \mathcal{N}(0,1)$ is the prior latent distribution, $\mathbf{I}$ indicates the object image.
%and $q(\mathbf{z} \,|\, \mathbf{I})$ is a learnable Gaussian distribution where $z$ is sampled from. 
$\mathbf{z}$ is sampled from $q(\mathbf{z} \,|\, \mathbf{I})$ using the reprameterization trick as in VAEs~\cite{kingma2013auto} for optimization.
%This loss is optimized same as in VAEs~\cite{kingma2013auto}. 
This loss prevents redundant deformations and yields cleaner shapes as shown in supplementary.% as shown in \cref{fig:visual_hull_ambiguity}(a).  

\vspace{0.15cm}
\noindent\textbf{Dense Shape Prior.} 
In the absence of Lidar supervision, the occupancy loss as a major shape cue is conceptually a shape-from-silhouette reconstruction, which suffers from the familiar visual hull ambiguity~\cite{cheung2005shape} (see supp. for illustration). %(Fig.\ref{fig:visual_hull_ambiguity}(b))
%, causing unstable shape deformation. 
Inspired by~\cite{yu2021plenoctrees}, 
%Hence, inspired by~\cite{yu2021plenoctrees}, 
we apply a dense shape prior to 
favor solid over empty space if both are possible solutions,
\begin{equation}
    \mathcal{L}_{dense} =  \frac{\sum_s^S  exp\big( \!-\!\Phi^{[\sigma]}(\mathbf{o}_s) \cdot d \, \big)}{S},
\end{equation}
where the prior is applied to S randomly sampled NOCS $\mathbf{o}_s$. $d\!=\!0.05$ is a hyper-parameter indicating an interval.
Note we only apply this prior in the absence of Lidar supervision.

\subsection{3D Object Localization}
\label{sec:localization}

\noindent\textbf{NOCS Prediction.} 
For learning NOCS, we enforce the consistency between the network-predicted NOCS and the NeRF-rendered ones from Eq.\eqref{eq:render_noc}, yielding the loss

\begin{equation}
   \mathcal{L}_{nocs} = \frac{\sum_{i \in \Omega_{fg}} || \mathbf{o}_i^{[pred]} - \mathbf{o}_i^{[render]} ||^2 } {\big|\Omega_{fg}\big|},
\end{equation}

\noindent where $\Omega_{fg}$ denotes all pixels within the ground-truth object mask.
%Note that the NOCS prediction branch is mainly a student who itself does not own shape information. 
%However, we found it beneficial to enforce bi-directional consistency by optimizing NeRF through this loss too, as it yields smoother shapes (see supplementary) and improves performance (\cref{seq:otherdesignchoice}). 
Finally, we regress a foreground object mask using a simple $L_2$ loss with the ground truth.

\vspace{0.15cm}
\noindent\textbf{PnP Solver and Score Prediction.} 
Combining the NOCS with the object size prediction from the base detector, we solve the PnP problem (\cref{eq:pnp_loss}) using Levenberg–Marquardt 
with an outlier-robust initialization scheme same as \cite{chen2022epro}. We use the predicted foreground probability along with a learned uncertainty map (detailed in supp.) as the per-pixel weight. This step yields the estimated 3D box. A complete 3D detector also needs to return confidence scores indicating the accuracy. So, we add a regression head with $L_2$ loss to predict the IoU between the estimated and ground truth 3D box similarly to~\cite{chen2021monorun}. This head takes as input the object feature map, the predicted NOCS map, and a Jacobian map. The Jacobian map is obtained by the partial derivatives of the PnP loss (\cref{eq:pnp_loss}) w.r.t. object center $\textbf{t}$ at each pixel, evaluated at the solved pose. Specifically, 
\begin{equation}
    \frac{\partial \mathbf{r}_i^{[rep]}} {\partial \mathbf{t}} = 
    \begin{bmatrix}
    \frac{\partial [\mathbf{r}_i^{[rep]}]_x} {\partial [\mathbf{t}]_x}
    & 0 & 
    \frac{\partial [\mathbf{r}_i^{[rep]}]_x} {\partial [\mathbf{t}]_z},
    \\
    0 & 
    \frac{\partial [\mathbf{r}_i^{[rep]}]_y} {\partial [\mathbf{t}]_y}
    & 
    \frac{\partial [\mathbf{r}_i^{[rep]}]_y} {\partial [\mathbf{t}]_z}
    \\
    0 & 0 & 0
    \end{bmatrix},
\end{equation}
where we flatten the non-zero elements as a feature vector for each pixel. The Jacobian measures the stability and correlates with the uncertainty~\cite{Wolfgang} of the solved pose, thus supplies the network with informative signals
%useful inductive bias 
to reason accuracy.
Both the NOCS and Jacobian map are detached here, with gradients backpropagated through the object feature map only. 
Finally, we multiply the predicted IoU score with the confidence score from the base detector as the final score. %same as \cite{peng2022did}. 

\vspace{0.15cm}
\noindent\textbf{Scale Fusion.} 
The metric scale of monocular localization is inherently ambiguous. In particular, the scale in PnP-based methods is solely determined by the object size prediction and could be unreliable~\cite{li2022diversity}. 
Here, we propose a simple yet effective method that fuses into our PnP method the direct object depth $d_{pred}$ additionally regressed by our base 3D detector.
%the direct object depth prediction $d_{pred}$ from the base 3D detector into our PnP-based method. 
We first update object size prediction $\mathbf{s}$ as 
\begin{equation}
    \mathbf{s}' =  \frac{d_{pred}+[\mathbf{t}]_z}{2\cdot[\mathbf{t}]_z}\,\mathbf{s}.
\end{equation}
$\mathbf{s}'$ inherently averages $\mathbf{s}$ with an optimal size $\frac{d_{pred}}{[\mathbf{t}]_z}\mathbf{s}$ that yields the prediction depth $d_{pred}$ in the current PnP problem. Next, we scale the translation estimation $\mathbf{t}$ in tandem 
\begin{equation}
    \mathbf{t}' = \frac{d_{pred}+[\mathbf{t}]_z}{2\cdot[\mathbf{t}]_z}\,\mathbf{t}.
\end{equation}
This retains the optimality of our pose with the reprojection error intact, and yet the scale is fused across object size prediction and object depth prediction for robustness. We have also attempted more sophisticated fusion as discussed in supplementary, but found this scheme to be sufficient. 

\begin{figure}[t!]
  \centering
  \vspace{-0.5em}
  \includegraphics[width=1.0\linewidth, trim = 0mm 103mm 148mm 0mm, clip]{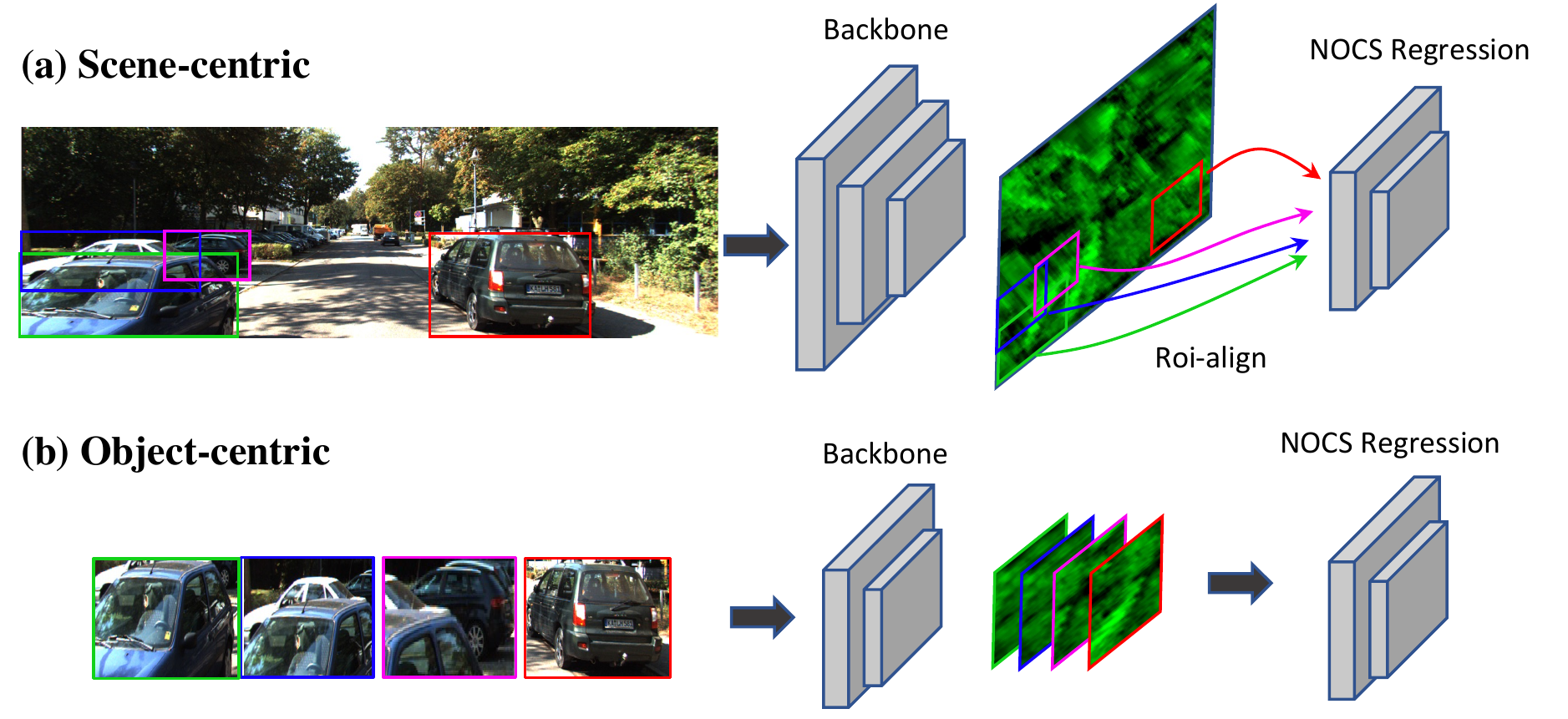}
  \centering
  \caption{Illustration of the scene-centric and object-centric training scheme that are shown to be an important design choice.}
  \label{fig:fullimagevscrop}
  \vspace{-0.5em}
\end{figure}

\begin{table*}[t!] 
    \centering
    \renewcommand{\arraystretch}{1.1}
    \resizebox{1.95\columnwidth}{!}{%
        \begin{tabular}{cccccccccccccccc}
\hline
\multicolumn{1}{c|}{\multirow{2}{*}{Class}} &
\multicolumn{1}{c|}{\multirow{2}{*}{Method}} &
\multicolumn{1}{c|}{\multirow{2}{*}{Venue}} &
\multicolumn{1}{c|}{\multirow{2}{*}{Cat.}} &
\multicolumn{3}{c|}{3D $AP_{40} $ - Test} & \multicolumn{3}{c||}{3D $AP_{40} $ - Val} & \multicolumn{3}{c|}{BEV $AP_{40} $ - Test}  & \multicolumn{3}{c}{BEV $AP_{40} $ - Val} \\

\cline{5-16} 
\multicolumn{1}{c|}{} &
\multicolumn{1}{c|}{} & \multicolumn{1}{c|}{} &
\multicolumn{1}{c|}{} & Easy & Moderate & \multicolumn{1}{c|}{Hard} & Easy & Moderate & \multicolumn{1}{c||}{Hard} & Easy & Moderate & \multicolumn{1}{c|}{Hard} & Easy & Moderate & Hard \\ \hline

\multicolumn{1}{c|}{\multirow{11}{*}{Regression}} &
\multicolumn{1}{c|}{Monodle~\cite{ma2021delving}} & 
\multicolumn{1}{c|}{CVPR21} &
\multicolumn{1}{c|}{E} &
17.23 & 12.26 & \multicolumn{1}{c|}{10.29}  &
17.45 & 13.66 & \multicolumn{1}{c||}{11.68} &
24.79 & 18.89 & \multicolumn{1}{c|}{16.00}  &
24.97 & 19.33 & 17.01 \\

\multicolumn{1}{c|}{} & \multicolumn{1}{c|}{MonoEF~\cite{zhou2021monocular}} & 
\multicolumn{1}{c|}{CVPR21} &
\multicolumn{1}{c|}{E} &
21.29& 13.87 & \multicolumn{1}{c|}{11.71}  &
-&  -& \multicolumn{1}{c||}{-} &
29.03& 19.70 & \multicolumn{1}{c|}{17.26}  &
-&  -& - \\

\multicolumn{1}{c|}{} & \multicolumn{1}{c|}{DDMP-3D~\cite{wang2021depth}} & \multicolumn{1}{c|}{CVPR21} &
\multicolumn{1}{c|}{E} &
19.71 & 12.78 & \multicolumn{1}{c|}{9.80}  &
\textcolor{MidnightBlue}{28.12} & \textcolor{MidnightBlue}{20.39} & \multicolumn{1}{c||}{\textcolor{MidnightBlue}{16.34}} &
28.08 & 17.89 & \multicolumn{1}{c|}{13.44}  &
-& - & -  \\ 

\multicolumn{1}{c|}{} & \multicolumn{1}{c|}{CaDDN~\cite{reading2021categorical}} & \multicolumn{1}{c|}{CVPR21} &
\multicolumn{1}{c|}{E} &
19.17& 13.41 & \multicolumn{1}{c|}{11.46}  &
{23.57} & 16.31 & \multicolumn{1}{c||}{13.84} &
27.94 & 18.91  &  \multicolumn{1}{c|}{17.19}  &
-& - & - \\ 
 
\multicolumn{1}{c|}{} & \multicolumn{1}{c|}{GrooMeD-NMS~\cite{kumar2021groomed}} & \multicolumn{1}{c|}{CVPR21} &
\multicolumn{1}{c|}{E} &
18.10 & 12.32  & \multicolumn{1}{c|}{9.65}  &
19.67 & 14.10 & \multicolumn{1}{c||}{10.47} &
26.19 & 18.27  & \multicolumn{1}{c|}{14.05}  &
27.38 & 19.75  & 15.92 \\ 
 
%\multicolumn{1}{c|}{} & \multicolumn{1}{c|}{DD3D~\cite{park2021pseudo}} & \multicolumn{1}{c|}{ICCV21} &
%\multicolumn{1}{c|}{E} &
% 23.19 & 16.87 & \multicolumn{1}{c|}{14.36}  &
% 32.35 & 23.41 & \multicolumn{1}{c||}{20.42}  &
% &  & \multicolumn{1}{c|}{} &
% &  &  \\ 

\multicolumn{1}{c|}{} & \multicolumn{1}{c|}{PCT~\cite{wang2021progressive}} &
\multicolumn{1}{c|}{NeurIPS21} &
\multicolumn{1}{c|}{E} &
21.00 & 13.37 & \multicolumn{1}{c|}{11.31}  &
\textcolor{MidnightBlue}{38.39} & \textcolor{MidnightBlue}{27.12} & \multicolumn{1}{c||}{\textcolor{MidnightBlue}{24.11}} &
29.65 & 19.03 & \multicolumn{1}{c|}{15.92}  &
\textcolor{MidnightBlue}{47.16} & \textcolor{MidnightBlue}{34.65} & \textcolor{MidnightBlue}{28.47} \\ 

\multicolumn{1}{c|}{} & \multicolumn{1}{c|}{MonoGround~\cite{qin2022monoground}} & \multicolumn{1}{c|}{CVPR22} &
\multicolumn{1}{c|}{E} &
21.37 & 14.36 & \multicolumn{1}{c|}{12.62}  &
25.24 & {18.69} & \multicolumn{1}{c||}{{15.58}} &
30.07& 20.47 & \multicolumn{1}{c|}{17.74}  &
32.68& 24.79 & 20.56  \\ 

\multicolumn{1}{c|}{} & \multicolumn{1}{c|}{HomoLoss~\cite{gu2022homography}} & \multicolumn{1}{c|}{CVPR22} &
\multicolumn{1}{c|}{E} &
21.75 & 14.94 & \multicolumn{1}{c|}{13.07}  &
23.04 & 16.89  & \multicolumn{1}{c||}{14.90} &
29.60& 20.68 & \multicolumn{1}{c|}{17.81}  &
31.04& 22.99 & 19.84 \\

\multicolumn{1}{c|}{} & \multicolumn{1}{c|}{MonoDTR~\cite{huang2022monodtr}} & \multicolumn{1}{c|}{CVPR22} &
\multicolumn{1}{c|}{E} &
 21.99 & 15.39 & \multicolumn{1}{c|}{12.73}  &
 24.52 & 18.57  & \multicolumn{1}{c||}{15.51} &
 28.59 & 20.38 & \multicolumn{1}{c|}{17.14}  &
 {33.33} & {25.35} & {21.68} \\ 

\multicolumn{1}{c|}{} & \multicolumn{1}{c|}{PseudoStereo~\cite{chen2022pseudo}} & \multicolumn{1}{c|}{CVPR22} &
\multicolumn{1}{c|}{E} &
 23.74& {17.74}  & \multicolumn{1}{c|}{{15.14}}  &
 \textcolor{MidnightBlue}{35.18} & \textcolor{MidnightBlue}{24.15} & \multicolumn{1}{c||}{\textcolor{MidnightBlue}{20.35}} &
 32.84 & {23.67}  & \multicolumn{1}{c|}{{20.64}}  &
 -& -  & - \\

\multicolumn{1}{c|}{} & \multicolumn{1}{c|}{DID-M3D~\cite{peng2022did}} & \multicolumn{1}{c|}{ECCV22} &
\multicolumn{1}{c|}{E} &
{24.40} & 16.29 & \multicolumn{1}{c|}{13.75}  &
25.38 & 17.07 & \multicolumn{1}{c||}{14.06} &
{32.95} & 22.76 & \multicolumn{1}{c|}{19.83}  &
33.91 & 24.00 & 19.52 \\ \hline

%\multicolumn{1}{c|}{MoGDE~\cite{mogde2022}} &
%\multicolumn{1}{c|}{NeurIPS22} &
%\multicolumn{1}{c|}{E} &
% &  & \multicolumn{1}{c|}{}  &
% &  & \multicolumn{1}{c||}{} &
% &  & \multicolumn{1}{c|}{}  &
%- & - & - \\ 

\multicolumn{1}{c|}{\multirow{4}{*}{Geometric}} & \multicolumn{1}{c|}{MonoRUn~\cite{chen2021monorun}} &
\multicolumn{1}{c|}{CVPR21} &
\multicolumn{1}{c|}{P} &
19.65 & 12.30 & \multicolumn{1}{c|}{10.58}  &
20.02 & 14.65 & \multicolumn{1}{c||}{12.61} &
27.94 & 17.34 & \multicolumn{1}{c|}{15.24}  &
- & - & - \\

%\multicolumn{1}{c|}{*EProPnP} & \multicolumn{1}{c|}{P} & - & - & \multicolumn{1}{c|}{-} &  &  & \multicolumn{1}{c||}{} & - & - & \multicolumn{1}{c|}{-} &  &  &  \\

\multicolumn{1}{c|}{} & \multicolumn{1}{c|}{MonoRCNN~\cite{shi2021geometry}} & % 
\multicolumn{1}{c|}{ICCV21} &
\multicolumn{1}{c|}{H} &
18.36 & 12.65 & \multicolumn{1}{c|}{10.03}  &
16.61 & 13.19 & \multicolumn{1}{c||}{10.65} &
25.48 & 18.11 & \multicolumn{1}{c|}{14.10}  &
25.29 & 19.22 & 15.30 \\ 

\multicolumn{1}{c|}{} & \multicolumn{1}{c|}{Autoshape~\cite{liu2021autoshape}} &
\multicolumn{1}{c|}{ICCV21} &
\multicolumn{1}{c|}{P} &
22.47 & 14.17 & \multicolumn{1}{c|}{11.36}  & 
20.09 & 14.65 & \multicolumn{1}{c||}{12.07} &
30.66 & 20.08 & \multicolumn{1}{c|}{15.95}  &
- & - & - \\

\multicolumn{1}{c|}{} & \multicolumn{1}{c|}{DCD~\cite{li2022densely}} &
\multicolumn{1}{c|}{ECCV22} &
\multicolumn{1}{c|}{P} &
23.81 & 15.90 & \multicolumn{1}{c|}{13.21}  &
23.94 & 17.38 & \multicolumn{1}{c||}{15.32} &
32.55 & 21.50 & \multicolumn{1}{c|}{18.25}  &
- & - & - \\

\if false
\multicolumn{1}{c|}{} & \multicolumn{1}{c|}{\textbf{\Ourstight-M (PnP)}} &
\multicolumn{1}{c|}{CVPR23} &
\multicolumn{1}{c|}{P} &
25.25 & 15.99 & \multicolumn{1}{c|}{13.27}  &
28.27 & 18.50 & \multicolumn{1}{c||}{15.69} &
\textbf{33.92} & 21.68 & \multicolumn{1}{c|}{18.27}  &
35.42 & 23.84 & 20.36 \\ 

\multicolumn{1}{c|}{} & \multicolumn{1}{c|}{\textbf{\Ourstight-MLC (PnP)}} &
\multicolumn{1}{c|}{CVPR23} &
\multicolumn{1}{c|}{P} &
\textbf{25.68} & \textbf{16.52} & \multicolumn{1}{c|}{\textbf{13.85}}  &
\textbf{28.46} & \textbf{18.68} & \multicolumn{1}{c||}{\textbf{16.22}} &
32.82 & \textbf{21.69} & \multicolumn{1}{c|}{\textbf{18.53}}  &
\textbf{35.65} & \textbf{24.55} & \textbf{20.98} \\ 
\fi
\hline % avg val model

\multicolumn{1}{c|}{\multirow{7}{*}{Hybrid}} & \multicolumn{1}{c|}{MonoFlex~\cite{zhang2021objects}} &
\multicolumn{1}{c|}{CVPR21} &
\multicolumn{1}{c|}{EH} &
19.94 & 13.89 & \multicolumn{1}{c|}{12.07}  &
23.64 & 17.51 & \multicolumn{1}{c||}{14.83} &
28.23 & 19.75 & \multicolumn{1}{c|}{16.89}  &
- & - & - \\

\multicolumn{1}{c|}{} & \multicolumn{1}{c|}{GUPNet~\cite{lu2021geometry}} &
\multicolumn{1}{c|}{ICCV21} &
\multicolumn{1}{c|}{EH} &
22.26 & 15.02  &  \multicolumn{1}{c|}{13.12} &
22.76 & 14.46  & \multicolumn{1}{c||}{13.72} &
30.29 & 21.19  &  \multicolumn{1}{c|}{18.20} &  
31.07 & 22.94  & 19.75\\

\multicolumn{1}{c|}{} & \multicolumn{1}{c|}{MonoJSG~\cite{liu2021autoshape}} &
\multicolumn{1}{c|}{CVPR22} &
\multicolumn{1}{c|}{EP} &
24.69& 16.14 & \multicolumn{1}{c|}{13.64}  & 
26.4 & 18.3 & \multicolumn{1}{c||}{15.4} &
32.59 & 21.26 & \multicolumn{1}{c|}{18.18}  &
- & - & - \\

\multicolumn{1}{c|}{} & \multicolumn{1}{c|}{MonoDDE~\cite{li2022diversity}} &
\multicolumn{1}{c|}{CVPR22} &
\multicolumn{1}{c|}{EHP} &
24.93 & 17.14 & \multicolumn{1}{c|}{15.10}  &
26.66 & 19.75 & \multicolumn{1}{c||}{16.72} &
33.58 & 23.46 & \multicolumn{1}{c|}{20.37}  &
35.51 & 26.48 & 23.07 \\

\multicolumn{1}{c|}{} & \multicolumn{1}{c|}{DEVIANT~\cite{kumar2022deviant}} & \multicolumn{1}{c|}{ECCV22} &
\multicolumn{1}{c|}{EH} &
 21.88 & 14.46 & \multicolumn{1}{c|}{11.89}  &
 24.63 & 16.54 & \multicolumn{1}{c||}{14.52} &
 29.65 & 20.44 & \multicolumn{1}{c|}{17.43}  &
 32.60 & 23.04 & 19.99 \\

\multicolumn{1}{c|}{} & \multicolumn{1}{c|}{\textbf{\Ourstight-M}} &
\multicolumn{1}{c|}{CVPR23} &
\multicolumn{1}{c|}{EP} &
\textbf{29.80} & \textbf{18.60} & \multicolumn{1}{c|}{\textbf{15.62}} &
\textbf{31.31} & \textbf{21.07} & \multicolumn{1}{c||}{\textbf{17.79}} &
\textbf{37.50} & \textbf{24.39} & \multicolumn{1}{c|}{\textbf{20.77}} &
 \textbf{39.26} & \textbf{26.91} & \textbf{23.69} \\

\multicolumn{1}{c|}{} & \multicolumn{1}{c|}{\textbf{\Ourstight-MLC}} &
\multicolumn{1}{c|}{CVPR23} &
\multicolumn{1}{c|}{EP} &
\textbf{29.89} & \textbf{18.94} & \multicolumn{1}{c|}{\textbf{15.90}} &
\textbf{31.24} & \textbf{21.01} & \multicolumn{1}{c||}{\textbf{17.70}}&
\textbf{37.27} & \textbf{24.49} & \multicolumn{1}{c|}{\textbf{20.89}} & \textbf{39.16} & \textbf{26.78} & \textbf{23.63}
 \\ \hline % avg val model

\multicolumn{16}{r}{\small {\Ourstight-M = Trained w/ Mask, \; \Ourstight-MLC = Trained w/ Mask+Lidar+LidarComp, \; \textbf{P} = PnP optimization and its variants, \; \textbf{E} = Direct depth estimation, \; \textbf{H} = Depth from height}}
\end{tabular}
    }
    \caption{\textbf{Comparisons with the state-of-the-arts in KITTI Benchmark,} using $AP_{40}$ with IoU$\geq$0.7 on test and validation set. Note that some methods use depth prediction from DORN~\cite{FuCVPR18-DORN} whose training data overlaps with the validation set as observed by~\cite{wang2020task,wang2019pseudo,chen2021monorun}, causing data leakage; these results are marked by \textcolor{MidnightBlue}{blue}. %The best APs within each class of methods and among all methods are in \textbf{bold} and \textcolor{\bestapcolor}{\textbf{red bold}}.
    }
    \label{table:benchmark}
\end{table*}

\subsection{Key Design Choices}
\label{sec:keydesignchoice}
\noindent\textbf{Visual Scope Matters.} 
We discuss a critical design choice among two common strategies in obtaining object-level feature for NOCS regression, as illustrated in \cref{fig:fullimagevscrop}.
One option (\eg \cite{zakharov2020autolabeling}) is to simply crop each object from the input image and regress NOCS from an \textit{object-centric} view. 
This is opposed to another \textit{scene-centric} strategy that takes the full image as network input and then crops the region-of-interest (RoI) from deep feature maps (\eg from a detection backbone) to regress NOCS (e.g.~\cite{chen2021monorun,lian2022monojsg}). The impact of this distinction on the NOCS learning and the subsequent 3D localization has not been well-understood. We conduct extensive experiments in \cref{seq:otherdesignchoice} to study this and observe the object-centric strategy to be significantly superior. 

Intuitively, we note that the scene-centric strategy retains rich context which is certainly essential for context-dependent tasks such as object depth estimation. However, object appearance alone has sufficient information to learn NOCS, as it is an intrinsic 3D property of the object. We conjecture that an object-centric view without scene context may enforce the network to learn NOCS the hard way -- a strict mapping from object appearance to NOCS. Conversely, rich context in scene-centric view may cause context bias~\cite{singh2020don},
%offer a shortcut \cite{geirhos2020shortcut}
allowing the network to rely on context instead of object appearance, which may hamper generalization with larger input variations. 
In \cref{seq:otherdesignchoice}, we observe that such an object-centric reasoning yields greater benefits for the challenging cases of distant or occluded objects.

\vspace{0.15cm}
\noindent\textbf{Implementation Details.} 
We combine all losses for joint training, with the weight of each term in supplementary. We apply a simple test-time augmentation by averaging inference on the flipped image. We use the code of~\cite{peng2022did} as base 3D detector. Detailed network architecture, computation efficiency, and more technical details are in supplementary.

\section{Experiments}

\subsection{Dataset and Evaluation Metrics}
\noindent \textbf{KITTI.} Following existing works we use KITTI-Object~\cite{geiger2012we} dataset as the main evaluation benchmark, focusing on the \textit{Car} category. This dataset contains a total of 7481/7518 training/test images. For ablation purpose the former is further split into 3716/3769 training and validation images~\cite{chen20153d}. Our framework uses the instance mask annotations provided by \cite{heylen2021monocinis}.
We use the standard evaluation metric $AP_{40}$~\cite{simonelli2019disentangling} - the average precision sampled at 40 recall positions in the precision-recall curve. The AP is computed for both 3D boxes and BEV boxes on the ground.
The objects are grouped into three difficulty levels - easy, moderate and hard.

\noindent \textbf{Waymo and NuScenes.} We follow \cite{kumar2022deviant,lian2022monojsg,wang2021progressive} to train and evaluate on Waymo~\cite{Sun_2020_CVPR} dataset using its front camera. In addition, we follow \cite{shi2021geometry,kumar2022deviant} to perform cross-dataset evaluation on NuScenes~\cite{caesar2020nuscenes}. These results are discussed in the supplementary material in the interest of space.

\begin{figure*}[]
    \centering
    \vspace{-0.2cm}
    \begin{tabular}{cccc}
    \subfloat[Impact of NeRF]
    {\label{fig:ablation_supervision}
    \includegraphics[width=0.5\columnwidth]{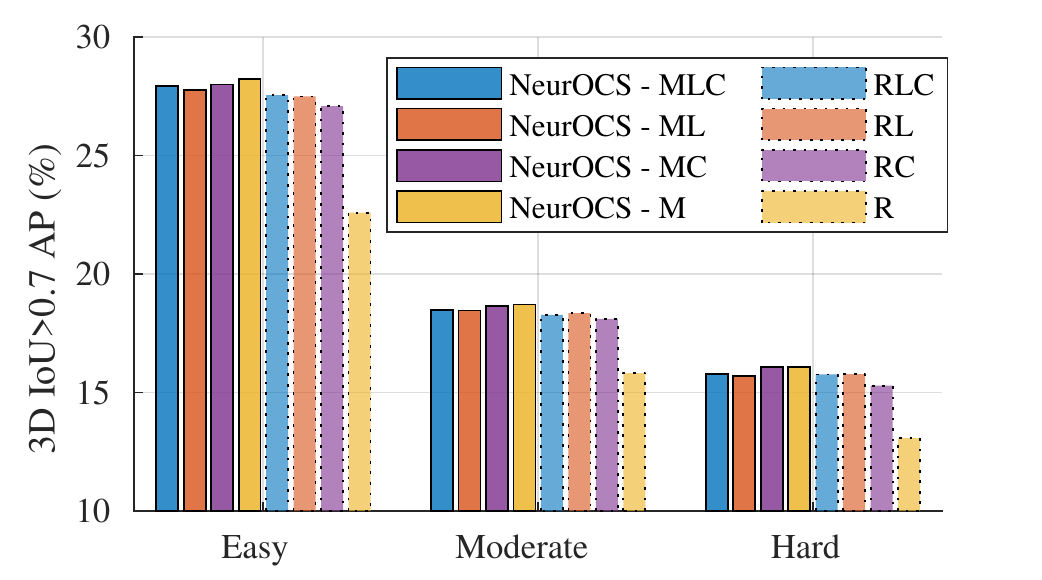}} 
    \!\!\!\!\!
    &
     \!\!\!\!\!
    \subfloat[Deformation Bases]
    {\label{fig:ablation_deformation}
    \includegraphics[width=0.5\columnwidth]{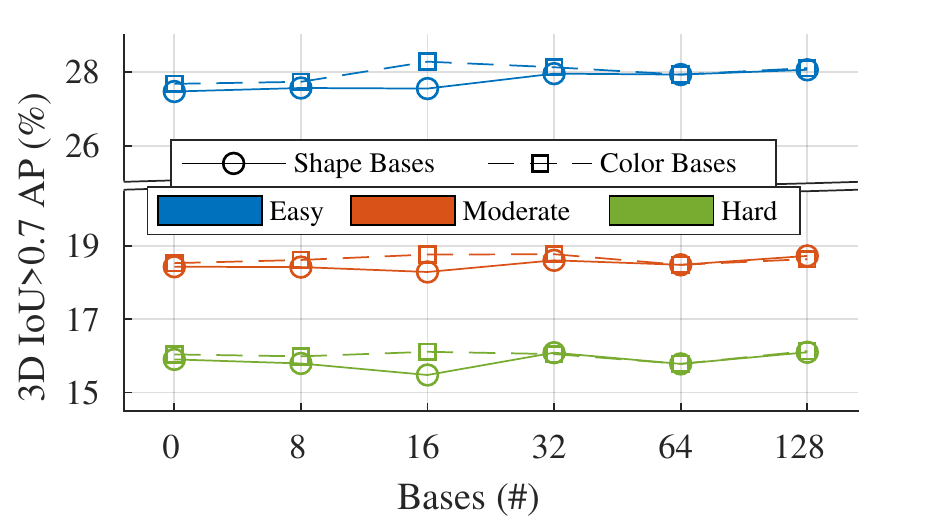}}
    \!\!\!\!\!
    & 
    \!\!\!\!\!
    \subfloat[Object-Centric v.s. Scene-Centric]
    {\label{fig:ablation_cropvsfull}
    \includegraphics[width=0.5\columnwidth]{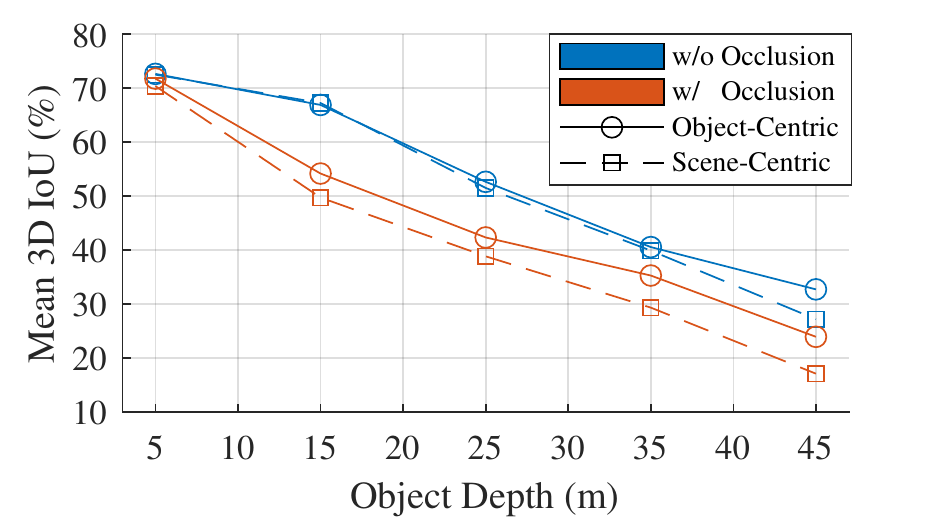}}
    \!\!\!\!\!
    & 
    \!\!\!\!\!
    \subfloat[Impact of Object-centric Focus]
    {\label{fig:ablation_shortcut}
    \includegraphics[width=0.5\columnwidth]{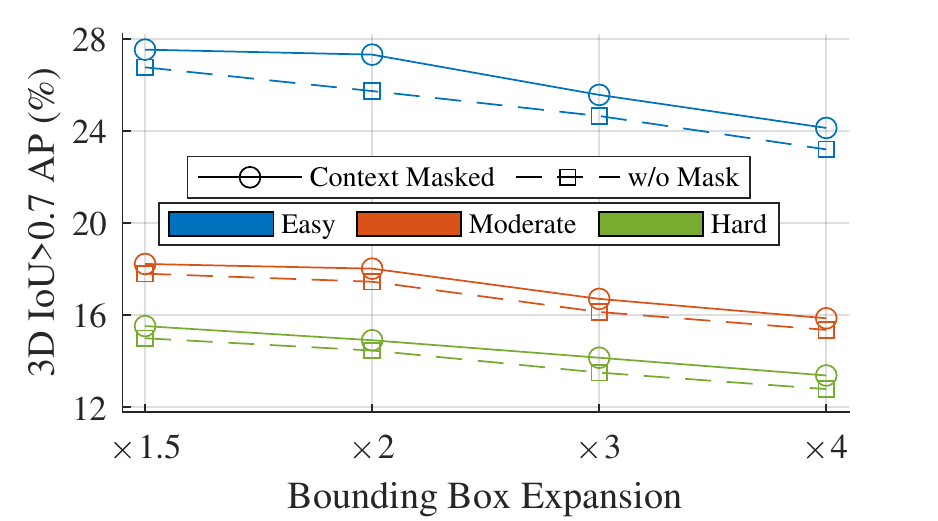}}
    \end{tabular}
    \caption{\textbf{Performance analysis } of (a) Impact of NeRF, (b) deformation bases, (b)(c) object-centric focus. The abbreviation look-up in \cref{fig:ablation_supervision} is \textbf{L}=Lidar, \textbf{C}=Lidar Completion, \textbf{M}=Mask, \textbf{R}=Reprojection Loss. All results are from the PnP solution without fusion.}
    %See text for details.}
    %\vspace{-0.5em}
    \label{fig:performancestudy}
\end{figure*}

\subsection{Evaluation on KITTI Benchmark}
\label{sec:kittibenchmak}
%As shown in \cref{table:benchmark}, we evaluate our method on KITTI-Object benchmark, both validation and test set, and compare with the state-of-the-art methods. 

\begin{figure}[t!]
    \centering
    \vspace{-0.3cm}
    \includegraphics[width=1.0\linewidth, trim = 0mm 82mm 112mm 0mm, clip]{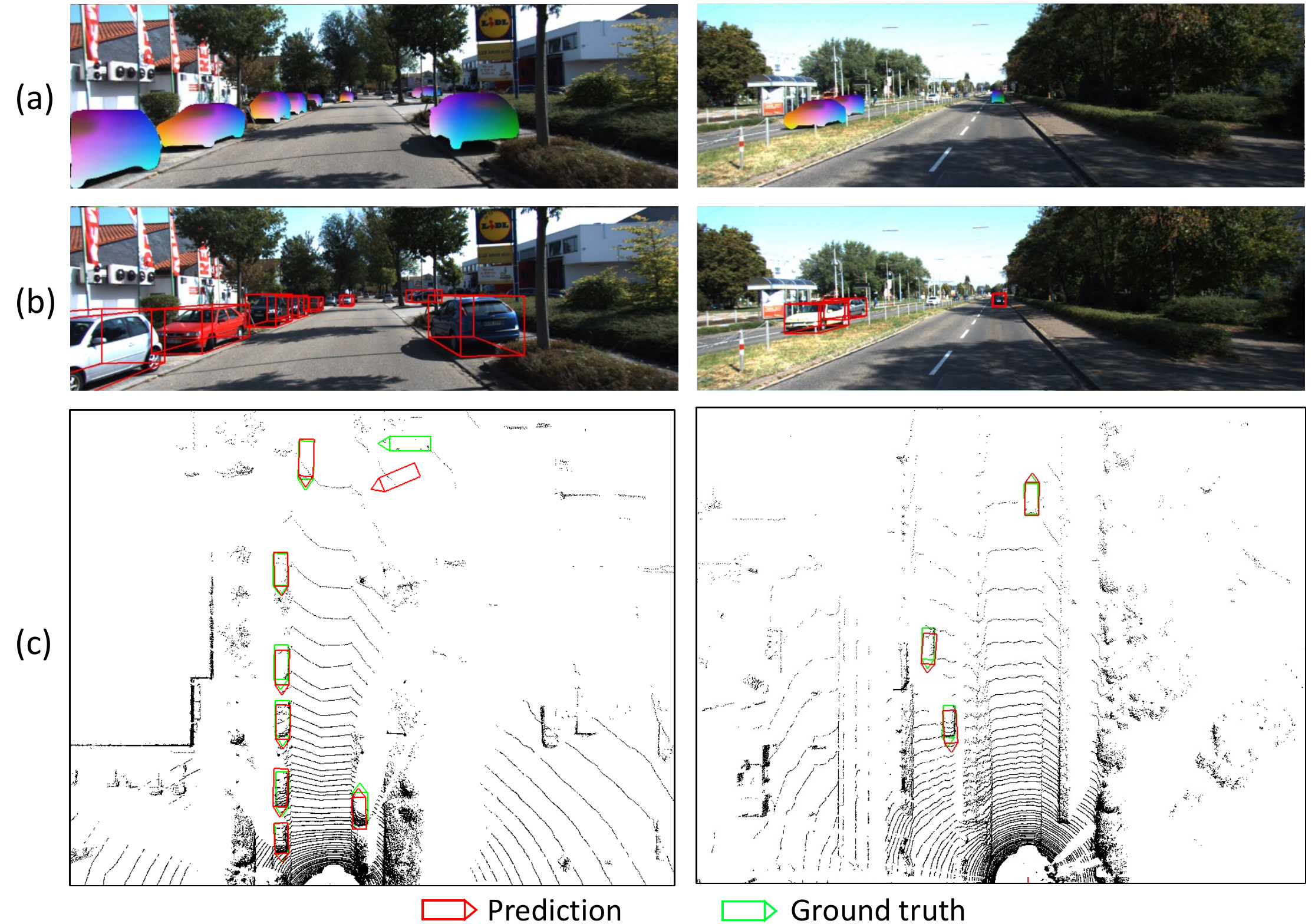}
    \vspace{-1.5em}
    \caption{\textbf{Qualitative examples} of our method with (a) predicted NOCS and (b) 3D boxes; (c) BEV boxes are also plotted on Lidar point cloud with the ground truth.}
    \label{fig:expe_qualitative}
    \vspace{-0.5em}
\end{figure}

We report the evaluation results in \cref{table:benchmark}. As discussed in \cref{sec:relatedwork}, the existing methods are grouped into three categories according to how they obtain object depth or location, 
%the most critical factor in localization~\cite{ma2021delving}. 
including direct depth regression, geometric methods, and hybrid methods.
The geometric cues may arise from PnP-like optimization with sparse or dense keypoints, as well as depth from height or edges.
%We report the results from \Ours both with PnP alone and with additional scale fusion with the direct depth regression from the base 3D detector~\cite{peng2022did}. 
We report the results from \Ours with scale fusion with the direct depth regression from the base 3D detector~\cite{peng2022did}. 
Also, we report results both when our shape model is trained with instance masks as the major shape cue without using Lidar (\Ourstight-M), and when Lidar and its completion are also applied for shape supervision (\Ourstight-MLC).
%Also note that we report here the result with auxiliary supervision with Lidar and its dense completion, while leaving other settings as ablation study to the next section. 
We do not compare here to the orthogonal works \cite{hong2022cross,peng2022lidar} that mainly rely on extra unlabeled data for improvements, leaving the discussion to supp. material.
As per common practice, the comparisons are primarily on the test set where the evaluation is done by KITTI servers using withheld ground truth, although we also provide results on validation set for a reference. 
%As can be seen, our PnP-based solution significantly outperforms existing methods that primarily rely on geometric reasoning as well. 
%In addition, it is complementary to direct object depth regression - its scale fusion with the base detector DID-M3D~\cite{peng2022did} results in further improvement, 
As can be seen, NeurOCS achieves state-of-the-art accuracy superior to existing methods with a large margin. 
In addition, \Ourstight-M yields strong performance close to or superior to \Ourstight-MLC. This demonstrates the power of our shape model that effectively translates mask annotations into NOCS supervision. We provide qualitative examples of our results in \cref{fig:expe_qualitative} to demonstrate its effectiveness; more results running on entire sequences are shown in supplementary.

\subsection{Analysis and Ablation Study}
Next, we conduct various analyses and ablation studies on validation set to understand the behavior of \Ourstight.

\subsubsection{The Role of NeRF}
\label{sec:nocssupervsion}
We study the impact of NeRF under different supervision.
\vspace{0.05cm}
\noindent \textbf{w/  NeRF.} In addition to  \Ourstight-M and \Ourstight-MLC discussed in \cref{sec:kittibenchmak}, we also evaluate \Ourstight-ML and \Ourstight-MC, that add Lidar or Lidar completion as supervision to \Ourstight-M.
%We denote our weakly-supervised shape learning as ``Ours-M", with instance masks as the shape cue without using Lidar. Ours with additional supervision from Lidar, Lidar completion, or both are respectively denoted as ``Ours-ML", ``Ours-MC", and ``Ours-MLC".

\vspace{0.05cm}
\noindent \textbf{w/o  NeRF.} Then, we remove NeRF and train the NOCS prediction branch directly with raw depth supervision, similarly to~\cite{chen2021monorun,lian2022monojsg}. Here, the counterpart to our mask-only supervision is the reprojection error loss proposed by~\cite{chen2021monorun}, denoted as ``R". Again, additional supervision from Lidar, Lidar completion, or both are denoted as ``RC", ``RL", and ``RLC".

We compare performance with the PnP-only solution to remove impact from fusion, as shown in \cref{fig:ablation_supervision}. First observe that ``\Ourstight-M" outperforms ``R" by a significant margin. This indicates the effectiveness of our method in leveraging instance masks as a weak supervision for NOCS learning even without Lidar. Conversely, reprojection errors impose weaker shape constraints as the depth ambiguity along the viewing ray persists, which does affect the downstream 3D localization. The usage of Lidar or its completion largely improves the performance when without using NeRF.
But the NeRF-based shape model results in superior accuracy by serving as a bridge between the raw Lidar data and the NOCS network that improves the supervision quality.
%Also worth noting here is an observation that direct regression with Lidar supervision alone yields poor accuracy; this is in accordance with the finding in~\cite{chen2021monorun}.

\begin{figure*}
    \centering
    \vspace{-1.0em}
    \includegraphics[width=1.7\columnwidth]{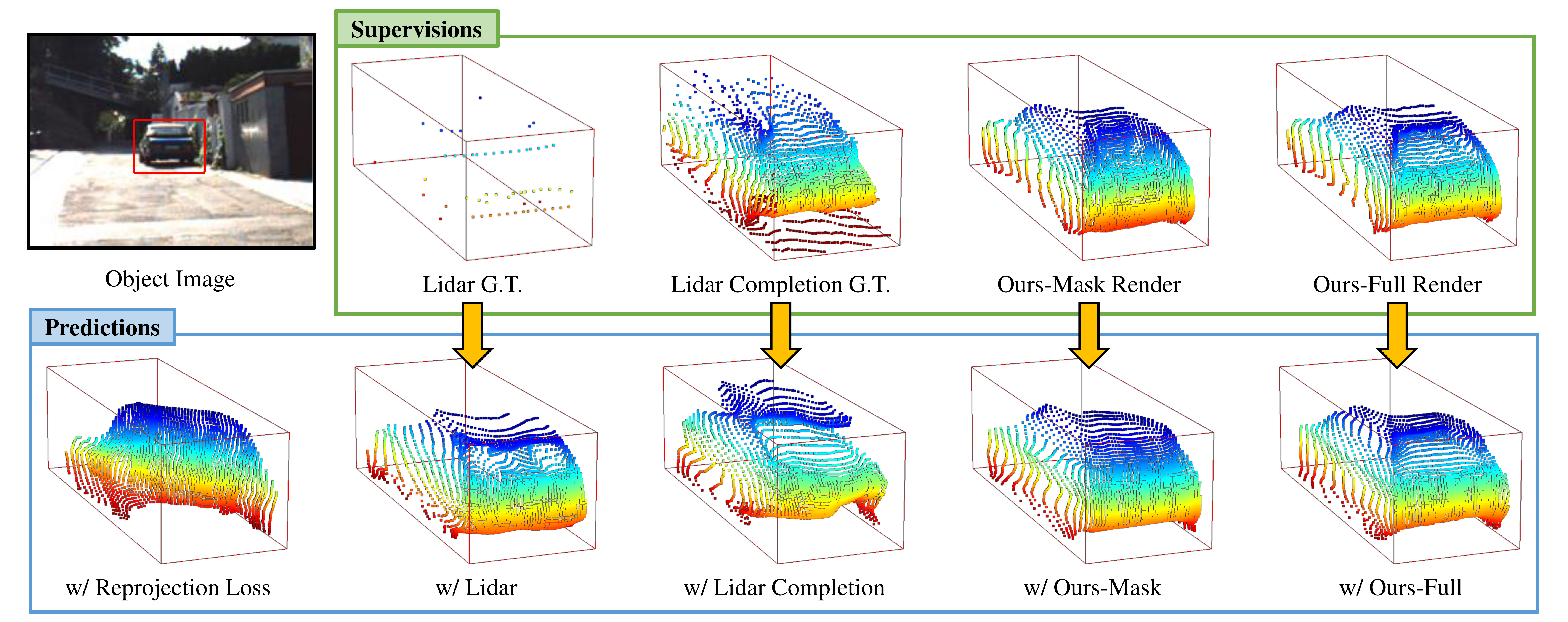}
    \caption{\textbf{Qualitative comparison of shapes learnt from different sources of supervision.} We visualize the NOCS by projecting them to point clouds with ground-truth object size. The point cloud is colored with Y-axis coordinate (height).}
    \label{fig:results_pc_back}
\end{figure*}

Next, we demonstrate qualitative examples of different supervisions in the form of point clouds, shown in \cref{fig:results_pc_back}.  The corresponding NOCS regression outputs under different supervisions are also visualized. One observes that our use of NeRF induces high-quality dense NOCS supervsion and hence prediction, in comparison to sparse Lidar and its completion. Remarkably, our weakly-supervised NOCS learning with instance masks yields far better shape than the self-supervised reprojection error loss. 

%\subsubsection{Local Crop vs. Global Context}

\subsubsection{Key Design Choices}
\label{seq:otherdesignchoice}
Next, we study the impact of various design choices in \cref{table:ablation} and \cref{fig:performancestudy}(b)-(d), using the validation set.
%based on \Ourstight-MLC as a default setting.

%reporting the performance both with PnP only and with additional fusion with~\cite{peng2022did}.
%Next, we study a number of important design choices listed in \cref{table:ablation}.

\begin{table}
    \centering
    \vspace{-0.3cm}
    \renewcommand{\arraystretch}{1.4}
    \resizebox{1.0\columnwidth}{!}{%
        \if false
\begin{tabular}{ll|ccc|ccc}
\hline
 & \multicolumn{1}{c|}{\multirow{2}{*}{}} & \multicolumn{3}{c|}{PnP Only} & \multicolumn{3}{c}{+ Fusion} \\ \cline{3-8} 
 & \multicolumn{1}{c|}{} & Easy & Moderate & Hard & Easy & Moderate & Hard \\ \hline
 & \textbf{Default} & \textbf{28.46} & \textbf{18.68} & \textbf{16.22} & \textbf{31.18} & \textbf{20.90} & \textbf{17.71} \\
w/o & Deform Reg. & -5.10 & -3.04 & -2.86 & -3.68 & -2.40 & -1.58 \\
w/o & Lidar & -0.44 & -0.07 & -0.52 & -0.44 & -0.30 & -0.55 \\
w/o & LiComp. & -0.32 & +0.19 & -0.05 & -0.14 & -0.12 & -0.36 \\
%w/o & Li. \& LiC. & -0.75 & -0.51 & -0.76 & -0.51 & -0.38 & -0.60 \\
w/o & Li. \& LiC. & -0.19 & -0.18 & -0.53 & -0.55 & -0.37 & -0.59 \\
%[-0.1891871  -0.17828394 -0.52482289] [-0.55297856 -0.36794892 -0.59160225]
w/o & Li. \& LiC. \& D.P. & -1.09 & -0.51 & -0.68 & -1.06 & -0.56 & -0.46 \\
w/o & DetScore & -2.56 & -0.97 & -0.37 & -2.48 & -0.85 & -0.35 \\
w/o & DetScore \& Jac. & -5.23 & -2.52 & -1.52 & -5.92 & -2.78 & -1.71 \\
w/o & Bi-Dir. NOCS  & -0.71 & -0.39 & -0.37 & -0.84 & -0.39 & -0.53 \\
w/o & T.T.A. & -0.81 & -0.37 & -0.29 & -0.24 & -0.20 & -0.36 \\
& Scene-centric & -4.18 & -2.11 & -2.59 & -3.07 & -1.73 & -1.39 \\
\hline
\end{tabular}
\fi

% \usepackage{multirow}

\begin{tabular}{ll|ccc|ccc}
\hline
 & \multicolumn{1}{c|}{\multirow{2}{*}{}} & \multicolumn{3}{c|}{PnP Only} & \multicolumn{3}{c}{+ Fusion} \\ \cline{3-8} 
 & \multicolumn{1}{c|}{} & Easy & Moderate & Hard & Easy & Moderate & Hard \\ \hline
 \multirow{4}{*}{Supervision} & NeurOCS-MLC & 27.92 & 18.49 & 15.78 & 31.24 &21.01 & 17.70  \\
 & NeurOCS-ML &  27.78 & 18.45 &15.70 & 31.07 & 20.94 & 17.72 \\
 & NeurOCS-MC &  28.00 & 18.63 & 16.06 & 31.27 & 21.08 & 17.79 \\
 & NeurOCS-M &  28.22 & 18.71 & 16.08 & 31.31 & 21.07 & 17.79 \\ \hline
 \multirow{7}{*}{NeurOCS-MLC} & w/o Deform Reg. & 26.33 & 17.68 & 14.71 & 29.69 & 20.27 & 17.06 \\ 
 & w/o DetScore&  23.48 & 16.84 & 14.94 &  27.20 & 19.54 & 17.06\\
 & w/o DetScore \& Jac. & 20.02 & 14.61  & 13.39  & 22.77 & 16.84 & 15.18 \\
 & w/o T.T.A. & 27.54 & 18.28 & 15.42 & 31.02 & 20.92 & 17.57 \\
& Scene-centric & 24.55 & 16.58 & 13.78 & 28.30 & 19.61 & 16.55 \\
& Directly regress deform. & 27.18 & 18.02 & 14.88  &  30.29 & 20.54 & 17.18 \\
& Off-the-shelf Mask & 27.84 & 18.56 & 15.60 & 30.86 & 20.87 & 17.63\\
\hline
\multirow{2}{*}{NeurOCS-M}& w/o Dense Prior &  28.17 & 18.76 & 16.08 & 31.18 & 21.00 & 17.79 \\ 
& Off-the-shelf Mask & 27.74 & 18.74 & 15.70 & 30.89 & 20.98 & 17.49\\
\hline
\end{tabular}

    }
    \caption{\textbf{Ablation study} of various design choices using $AP_{3D}$. 
    %Numbers indicate the drop of 3D AP with respect to the first row.}
    }
    \label{table:ablation}
    \vspace{-0.5em}
\end{table}

\vspace{0.1cm}
\noindent \textbf{Supervision sources.}
We first report in \cref{table:ablation} the results under different sources of supervision, including NeurOCS-MLC, NeurOCS-ML, NeurOCS-MC, and NeurOCS-M, where ``M", ``L" and ``C" respectively denote supervision from instance mask, Lidar, and Lidar completion. We observe competitive performance across all the four settings, indicating the robustness of our method, and importantly the capability of our shape model in effectively translating instance mask into high-quality NOCS supervision. In addition, our PnP-based solution significantly outperforms existing geometric methods listed in \cref{table:benchmark} that primarily rely on geometric reasoning as well. Furthermore, we note that the fusion of PnP with the scale from direct depth regression consistently improves the performance, indicating their complementary nature. 
%Similar to discussions in \cref{sec:nocssupervsion}, we remove Lidar or/and its completion, denoted as ``w/o Lidar",``w/o LiComp." and ``w/o Li. \& LiC.". One observes only slight AP degradation, for both PnP only and with scale fusion. 
%This indicates that our weakly supervised shape learning  is already performant.
%This indicates the capability of our shape model in effectively translating instance mask into high-quality NOCS supervision even without Lidar.

\vspace{0.1cm}
\noindent \textbf{KL regularization.} We study the impact of the KL regularization loss on the shape basis coefficients and find it to be important. As shown in \cref{table:ablation} ``w/o Deform Reg.", removing this loss from NeurOCS-MLC leads to a large drop in accuracy.  This implies server ambiguities in the shape modeling despite the low-rank structure in the latent grid, due to the challenging conditions in real driving scenes.

\noindent \textbf{Deformation bases.} We study the performance with varying number of deformation bases for both shape and color space, as shown in \cref{fig:ablation_deformation}. 
%Note that zero basis implies using solely a canonical mean latent grid. 
We observe that our NeRF-rendered NOCS leads to good localization performance even with just the canonical shape model, \ie using zero basis. However, the presence of deformation bases improves the accuracy by accounting for instance-level variations. In practice, we use 64 bases as a trade-off of accuracy and computation.
%Empirically, we observe the overall best performance with 64 bases.

\noindent \textbf{Directly regressing deformations.} Instead of decomposing the latent grid of NeRF as the shared canonical one plus per-instance linear deformation with shared basis, an alternative way is to let the network directly regress the latent grid for each instance independently without any constraints. We observe that this yields degraded performance as shown in \cref{table:ablation}, which indicates the benefits of the low-rank inductive bias in categorical shape learning.

\vspace{0.1cm}
\noindent \textbf {Visual scope.}
%\label{sec:expe_visualscope}
We study the behavior of the object-centric and scene-centric training strategies (\cref{sec:localization}).
%(1) cropping object patches from the raw image (2) cropping object patches from the feature map returned from a backbone network. 
%Both of their feature maps are resized to the same dimension before feeding to NOCS regression networks,
%but ``crop-featmap" contains rich context information.
We follow~\cite{chen2021monorun} to use ResNet101 with FPN~\cite{lin2017feature} to handle object scales in scene-centric scheme. Similar results with DLA~\cite{yu2018deep} are in supplementary. We report the AP metric for the scene-centric case in \cref{table:ablation}, which shows the default object-centric training is largely superior. The same is observed in supplementary when training with Lidar without NeRF.
To understand this benefit, we analyze per-instance 3D IoU w.r.t the ground truth 3D boxes.
Specifically, we group all recalled objects according to their ground truth depth, separate objects under occlusions (or truncations) from fully visible ones, and
then plot the mean 3D IoU for each group against object depth.
The results with direct Lidar training is shown
in \cref{fig:ablation_cropvsfull}. 
We observe that the object-centric training benefits more to distant and occluded objects -- regardless of the occlusions, the performance gap between the two schemes increases as depth increases; and evidently the performance gap is much larger for objects under occlusions. The performance gap remains even with larger network capacity (detailed in supp.)
%In supplementary, we observe the same even with larger network capacity in scene-centric training.

We further study the benefit of object-centric focus with its slight variant -- enlarging the detection boxes before cropping to include more scene context. We evaluate two settings: (a) feeding the object patches to networks as is; (b) masking out the additional context before doing (a). The additional context reduces the degree of object-centric focus, and in this way we isolate its effect. Results with direct Lidar training is shown in \cref{fig:ablation_shortcut}, where context masking-out yields higher accuracy, with a larger gap as the boxes expand more, 
%We observed the same even when they use the same object mask prediction, as shown in supplementary.
demonstrating the benefits of a greater object-centric focus.

%\vspace{0.05cm}
%\noindent \textbf{Bi-directional consistency.} We observe empirically that the bi-directional consistency in $L_{nocs}$ facilitates the  learning of NeRF, yielding smoother shape model (see supplementary) and AP gains, as shown in \cref{table:ablation} ``w/o Bi-Dir. NOCS".

\vspace{0.05cm}
\noindent \textbf{Detection score and Jacobian map.} We first remove the detection score from the base 3D detector (\cref{table:ablation} ``w/o DetScore"), and then additionally remove the Jacobian map from our score branch (``w/o DetScore \& Jac."). The results show our PnP score itself is meaningful, and the Jacobian map contributes positively to the score prediction.

\vspace{0.05cm}
\noindent \textbf{Test-time augmentation (T.T.A.).} We note that accuracy improvement (\cref{table:ablation} ``w/o T.T.A.") is brought by  T.T.A.

\vspace{0.1cm}
\noindent \textbf{Dense shape prior.} In the absence of Lidar, removing the dense prior from NeurOCS-M results in a drop (\cref{table:ablation} ``w/o Dense Prior"), indicating the dense prior may mitigate the visual hull ambiguity in single-view shape learning.

\vspace{0.05cm}
\noindent \textbf{Off-the-shelf instance masks.} Instead of the ground truth masks, we adopt the mask prediction from an off-the-shelf pretrained Mask R-CNN~\cite{he2017mask} model. This obviates the need for additional mask annotations in KITTI. We report in \cref{table:ablation} the accuracy for  NeurOCS-MLC and NeurOCS-M. While the performance drops as expected, it remains strong and outperforms existing methods shown in \cref{table:benchmark}. This demonstrates its robustness with respect to instance masks.

\section{Limitations and Conclusion}
 %In summary, our work proposes a new method towards NOCS-based 3D localization. It achieves new state-of-the-art performance driven by learning NOCS from categorical NeRF supervision in a context-independent way, as well as several important design choices. Similar to existing works with auto-labeling~\cite{liu2021autoshape}, our framework requires instance mask annotations despite its effectiveness. However, the labeling cost may be constantly reduced in the future with automation by the rapid advancement of networks in understanding semantics, e.g. panoptic segmentation. 
 %Further exploration in this direction remains our future work.
 %Finally, we envision that our study on the benefits of object-centric focus may carry implications for other tasks in computer vision. 
 
 In this work, we propose a 3D localization framework that rests on NOCS-based object pose estimation but leverages NeRF to address its key challenge -- the lack of supervision in real driving scenes. We learn category-level neural shape models to provide high-quality NOCS supervision. Driven by crucial design choices for effective NOCS learning and ambiguity handling, our method yields new state-of-the-art performance. One limitation of our method lies in its reliance on a base 3D detector and its object size prediction. Also, vehicles with irregular shape may cause challenges to our NOCS prediction. These remain interesting problems to explore in the future.
 %Our training currently requires instance mask annotations, which remains a limitation worthy to address in future work. 
 Further, we envision our work to inspire more efforts towards further unleashing the potential of differentiable rendering for 3D object localization.

%%%%%%%%% REFERENCES
{\small
\bibliographystyle{ieee_fullname}
\bibliography{egbib}
}

\end{document}